\documentclass[letterpaper]{article}
\usepackage{aaai2027}
\usepackage{times}
\usepackage{helvet}
\usepackage{courier}
\usepackage[hyphens]{url}
\usepackage{graphicx}
\usepackage{amsmath}
\usepackage{amssymb}
\usepackage{booktabs}
\usepackage{newfloat}
\usepackage{caption}
\usepackage{algpseudocode}
\usepackage{array}
\usepackage{xcolor}
\definecolor{dgreen}{HTML}{1A7F37}
\definecolor{dred}{HTML}{C0392B}
\newcommand{\dg}[1]{{\scriptsize\textcolor{dgreen}{$+#1$}}}
\newcommand{\dr}[1]{{\scriptsize\textcolor{dred}{$-#1$}}}
\algrenewcommand\algorithmiccomment[1]{\hfill\(\triangleright\)~\emph{#1}}
\title{BONSAI: Evolvability-Guided Tree Search over Skills}

\author{
    Yash Priya Shastri,
    Anand Eswaran,
    Adnan Qidwai,
    Pankaj Thorat,
    Sachin Joshi
}
\affiliations{
    IBM Research\\
    \{ypshastri@ibm.com, anand.eswaran@ibm.com, Adnan.Qidwai1@ibm.com,
      pankaj.thorat@ibm.com, jsachind@in.ibm.com\}
}

\nocopyright

\begin{document}
\maketitle
\pagestyle{plain}
\thispagestyle{plain}
\raggedbottom

\begin{abstract}
A \textbf{skill} is a natural-language document that steers a frozen agent whose weights cannot be
updated, so any capability the agent lacks must be supplied in prose. Optimising a skill is therefore
optimising text against a score, and the standard recipe, which keeps any edit that raises a held-out
score, is blind in a specific way: a single score cannot tell a document perched on a narrow, overfit
spike from one resting on a broad plateau, even though only the second can still be improved. We
introduce \textbf{BONSAI}, a novel skill-optimisation framework that steers instead by
\textbf{evolvability}, the capacity of a region of document-space to keep producing viable variation
under further mutation, a property biology treats as separate from present fitness. BONSAI grows skills as a Monte-Carlo search tree in which every child document is a \emph{mutation} of its parent, and descends it under an upper-confidence selection rule whose exploitation term blends a skill's own fitness with the fitness of its mutational neighbourhood. Because every child is a mutation, the mean score recorded beneath a node estimates that neighbourhood's evolvability at no extra cost, so the rule concentrates budget on regions that keep improving while its exploration term keeps a currently weak branch in contention. BONSAI ships the single best-scoring document it finds, at no cost beyond the accept-if-better loop it replaces. With a frozen
30B agent and averaged over three benchmarks, BONSAI lifts held-out accuracy over the skill-free agent
by $\mathbf{23.13}$ points and improves on two budget-matched baselines, GEPA and SkillOpt, by
$\mathbf{3.87}$ and $\mathbf{3.97}$ points respectively.
\end{abstract}

\begin{figure*}[t]
\centering
\includegraphics[width=0.99\textwidth]{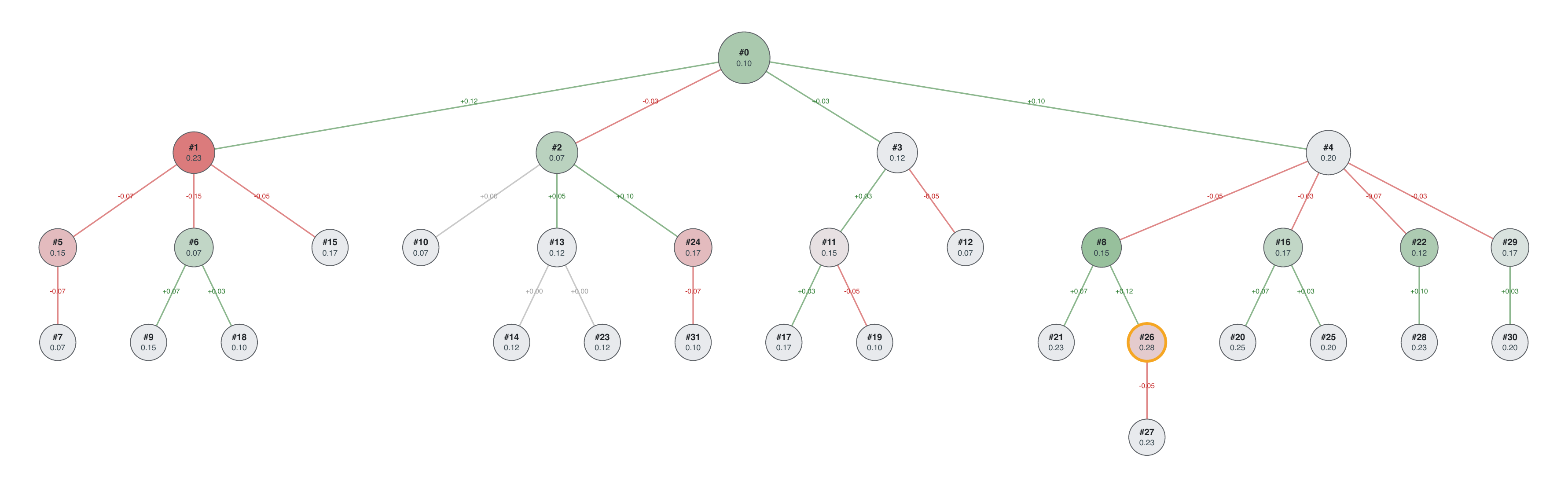}
\caption{A complete BONSAI search tree, the SpreadsheetBench run at seed 42: 32 skill documents
grown from a single seed document over 131 mutation proposals. Fill encodes brittleness $\sigma$ (green evolvable, red brittle), circle area encodes the
number of value samples $m(n)$, edge labels give the score change $\Delta v$ against the parent, and
the gold ring marks the shipped skill. Budget concentrates in a few evolvable lineages.}
\label{fig:tree}
\end{figure*}

\section{Introduction}

A frozen agent cannot learn from experience: its weights are fixed, so it cannot be trained further.
Whatever it is to do well must instead be told to it, in text. We call that text a \textbf{skill}: a
document placed in front of the model that guides how it approaches a family of tasks. A skill is
less a prompt template than a short field manual, stating which library to reach for, which edge
cases cause failures, and what to verify before returning an answer. Because the weights never move,
the skill is the only object an optimiser can touch, and every point of accuracy must be bought with
prose.

The natural way to improve a skill is to edit and test: show a model where the agent failed, let it
rewrite the document, and keep the rewrite when a score on held-out tasks rises
\cite{zhou2023ape,yang2024opro,khattab2024dspy,agrawal2025gepa,skillopt2026}. Several strong systems
are built on this loop, which nonetheless carries a blind spot: a validation score is one number over
a finite task set, so two documents that score alike may be quite different objects, one resting on a
broad plateau that further edits keep improving, the other on a narrow spike the next edit displaces.
The score alone cannot tell them apart, and a spike is a dead end, since any edit that repairs one
remaining failure tends to break something the document already handled.

The distinction is familiar elsewhere. Optimisation theory contrasts flat minima with sharp ones and
expects the flat ones to generalise better \cite{hochreiter1997flat,foret2021sam}, and biology treats
\emph{evolvability}, the ability of a lineage to keep producing useful variants, as a property
separate from present fitness \cite{wagner1996evolvability}. What has been missing for skills is a way
to measure this property that does not cost more than the search it guides.

\paragraph{Contributions:} First, we identify robustness under mutation, that is, evolvability, as
the property a skill optimiser should steer by, and we give a measurement of it that costs no extra model calls. Second, we turn that measurement into a search. Growing skills as a tree whose every child is a mutation makes the mean fitness beneath a node a measure of its region's evolvability, and we steer an upper-confidence tree search \cite{kocsis2006uct,browne2012mcts} by that measure, with a selection rule that blends a skill's own fitness with its neighbourhood's. Budget flows to regions where mutation keeps yielding improvement, while an exploration term revisits a currently weak neighbourhood before it is written off. Third, on SpreadsheetBench, at an equal and measured budget, BONSAI improves on the seed document it starts from by $5.71$ accuracy points on held-out tasks and
exceeds the strongest budget-matched baseline by $2.14$. The same ordering holds on SearchQA and
LiveMathematicianBench. Shipping stays separate from steering: we deploy the plain best-scoring document, for a
reason given below.

\section{Method}

\subsection{Skills, mutations, and the tree}

Two models play distinct roles, and neither is trained. The \textbf{performer} is the frozen agent:
it reads the skill, attempts a task, and is scored automatically. The \textbf{optimizer} is a second
model that never attempts tasks itself. It reads a small number of the performer's scored attempts,
each of which carries the task, the answer the performer produced, and the reason that answer was
judged incorrect, and it returns a rewritten skill document. The data is split once into a
\emph{train} set (from which small batches of tasks are drawn to prompt each rewrite), a
\emph{validation} set (used to score whole skills during the search), and a \emph{test} set (touched
once, at the very end).

\begin{figure*}[t]
\centering
\includegraphics[width=0.92\textwidth]{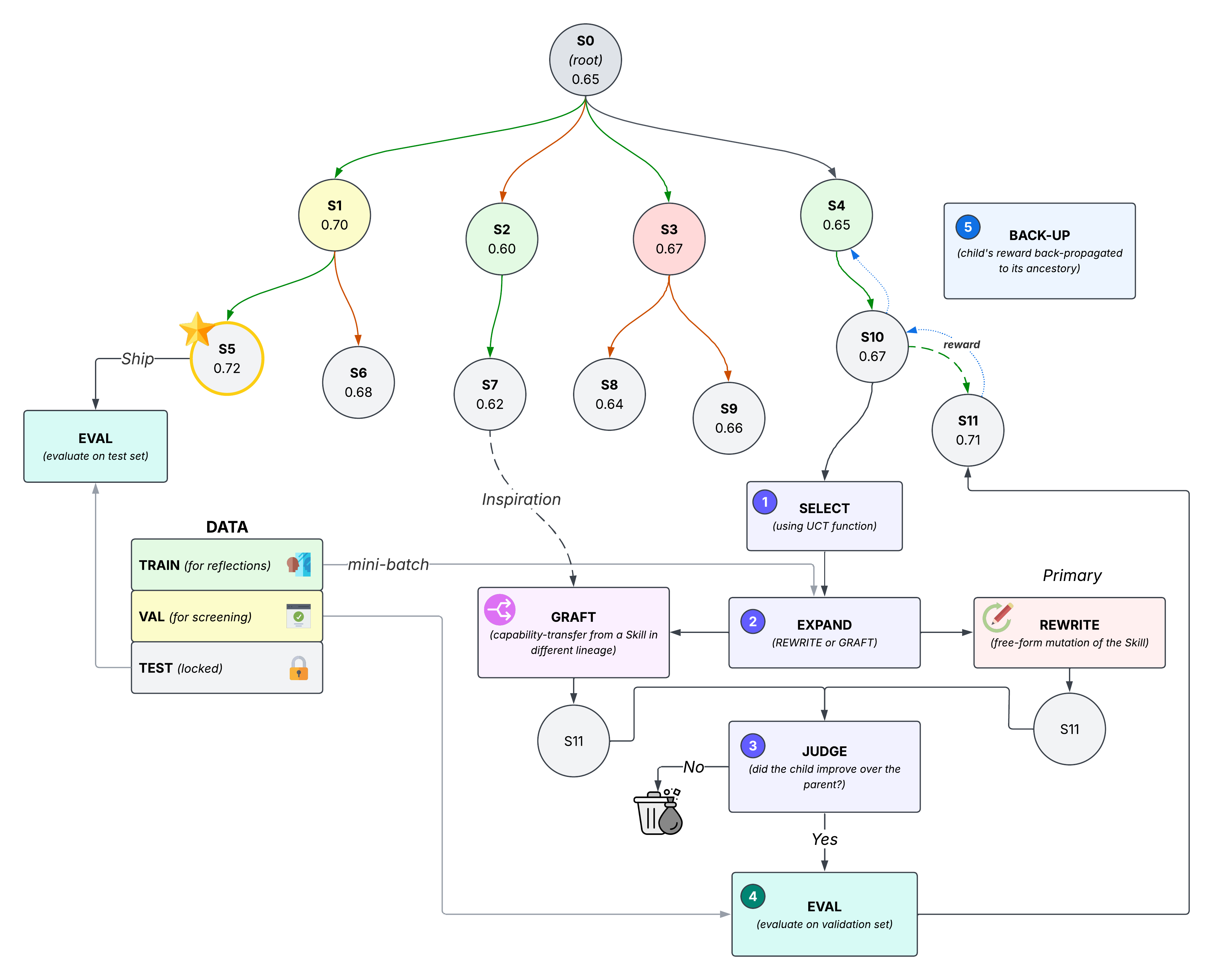}
\caption{BONSAI Workflow. The loop \textbf{selects} a node by rule~\eqref{eq:uct},
\textbf{expands} it by a free-form \emph{rewrite} (or, optionally, \emph{graft}), \textbf{judges} the
child against its parent on a training minibatch~\eqref{eq:accept}, \textbf{evaluates} an accepted
child on validation, and \textbf{backs} the result up the ancestry~\eqref{eq:backup}, a rejected child backs up a visit only~\eqref{eq:backup-reject}. The train / validation / test splits feed reflection,
screening, and the single final test. The gold-ringed node is the one shipped.}
\label{fig:pipe}
\end{figure*}

The search arranges skills into a tree (Figure~\ref{fig:pipe}). The root is the fixed seed document,
and an edge from node $n$ to node $c$ means that \emph{$c$ was produced by mutating $n$}, that is, by
one reflective rewrite. This single construction choice is what the remainder of the method rests
upon: it converts an unordered collection of candidate documents into a space with
neighbourhood structure, and a neighbourhood is something that can be measured.

\subsection{Evolvability}

Write $v(n)$ for the \textbf{fitness} of the document at node $n$, its validation score, the fraction
of held-out tasks it solves. Fitness is a point estimate, and two documents that score alike can
differ sharply in what surrounds them: one may sit on a broad plateau, where nearby rewrites score
just as well, the other on a narrow peak, from which the next edit falls away. What separates them is
a property of the terrain, not of the score itself.

We define \textbf{evolvability} as the expected fitness of a
skill's mutational neighbourhood: for a mutation operator that maps a skill to a child, the
evolvability of a skill $s$ is $\epsilon(s) = \mathbb{E}[v(s')]$, the expectation taken over the
children $s'$ mutation produces from $s$, and, transitively, over the region of skill-space reachable
from $s$ by repeated mutation. Evolvability thus belongs to a region rather than to a single document,
reporting not how a skill performs today but how well its future is likely to perform.

\begin{figure}[t]
\centering
\includegraphics[width=0.9\columnwidth]{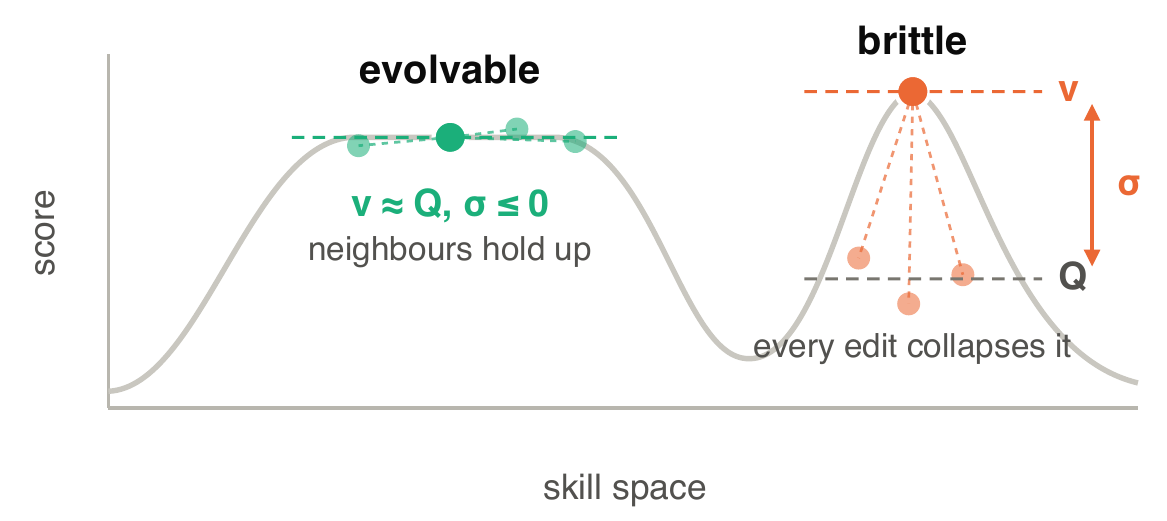}
\caption{Two skills with similar scores and opposite futures. The gap $\sigma = v - Q$ distinguishes a
brittle peak, whose neighbourhood scores far lower, from an evolvable region, whose neighbourhood
holds up.}
\label{fig:evolv}
\end{figure}

This region is unbounded, so $\epsilon$ cannot be read off directly, but the search tree gives a free,
self-improving estimator of it. Let $\mathcal{L}(n) = \{n\} \cup \mathrm{desc}(n)$ be node $n$'s
\emph{lineage}, the node together with every document later grown from it, and let $m(n)$ count the
lineage members scored so far. Every descendant was reached from $n$ by mutation, so the lineage's
mean fitness estimates $\epsilon(n)$. We call it the node's \textbf{evolvability value},
\begin{equation}
Q(n) \;=\; \frac{1}{m(n)} \sum_{s \in \mathcal{L}(n)} v(s)
\label{eq:Q}
\end{equation}
$Q(n)$ is a reading of fitness taken after perturbation, $v(n)$ the reading taken before it. The gap
between the two,
\begin{equation}
\sigma(n) \;=\; v(n) - Q(n)
\label{eq:sharp}
\end{equation}
we call \textbf{brittleness}. A large positive $\sigma$ marks a brittle, overfit peak. A $\sigma$ at or
below zero marks a document typical of, or even bettered by, a strong neighbourhood, the signature of
an evolvable region (Figure~\ref{fig:evolv}).

Two properties make $Q$ worth having. First, it is \emph{free}: every term in \eqref{eq:Q}
is a fitness value the search already paid for, so no extra calls are spent measuring it. Second, it
\emph{sharpens itself}: each expansion beneath $n$ adds a sample to $Q(n)$, so the nodes probed most
heavily are exactly those whose evolvability estimate becomes best resolved.

\subsection{The search}

BONSAI is a Monte-Carlo tree search whose selection rule is built to steer by evolvability. Each iteration selects a node, expands it by one mutation, judges the result, and propagates that
result back up the tree (Figure~\ref{fig:pipe}).

\paragraph{Selection:} Starting at the root, the search descends, taking at each level the child that
maximises an upper-confidence score,
\begin{equation}
U(s) \;=\; v(s) + \lambda\bigl(Q(s) - v(s)\bigr) + c\,\sqrt{\frac{\ln N(p)}{N(s)}}
\label{eq:uct}
\end{equation}
where $p$ is the parent of $s$ and $N(\cdot)$ counts how many times a node has been visited. At
$\lambda = 1$, which we use throughout, the exploitation term is exactly $Q(s)$, so evolvability leads
the search. At $\lambda = 0$ the rule reduces to plain fitness. The exploitation term is rescaled to
$[0,1]$ by the smallest and largest fitness recorded anywhere in the tree, a running normalisation
adapted from MuZero \cite{schrittwieser2020muzero}. This is what makes the exploration constant $c$
scale-free: one value behaves the same whether a benchmark's scores cluster near $10$\% or near $90$\%,
whereas on the raw scale $c$ would have to be retuned per benchmark to keep exploration and
exploitation in balance. The third term is the standard exploration bonus: a node whose neighbourhood
currently appears weak still draws visits, so that an early verdict can be revised rather than
standing.

\paragraph{Expansion and acceptance:} The optimizer proposes one child document, and it is kept only
if it strictly improves on the same batch of training tasks the optimizer was shown,
\begin{equation}
\sum_{i \in B} g_i(c) \;>\; \sum_{i \in B} g_i(n)
\label{eq:accept}
\end{equation}
where $g_i$ is the per-task score and $B$ is the batch. Judging the proposal on the batch that
produced it keeps the acceptance test aligned with the evidence behind the edit. An accepted child is
then scored on the full validation split to obtain its fitness $v(c)$.

\paragraph{Backup:} An accepted child sends its value up the ancestry. For every ancestor $a$,
\begin{equation}
N(a) \mathrel{+}= 1, \quad m(a) \mathrel{+}= 1, \quad W(a) \mathrel{+}= v(c)
\label{eq:backup}
\end{equation}
where $W(a)$ accumulates the backed-up scores and $Q(a) = W(a)/m(a)$. A rejected mutation produces no
document and no score, so it backs up a \textbf{visit only},
\begin{equation}
N(a) \mathrel{+}= 1, \quad m(a) \text{ and } W(a) \text{ unchanged}
\label{eq:backup-reject}
\end{equation}
Keeping the two counters apart matters more than it may appear. A failure indicates \emph{where not to look}. It is not a sample of a region's quality. Rule \eqref{eq:backup-reject} spends budget and
decays the exploration term in \eqref{eq:uct}, so the search moves on, while $Q$ is left untouched.
Conflating the counters would allow a run of failed rewrites to depress the estimate of a region that
was never shown to be worse.

\paragraph{Widening:} A node may hold only so many children, and the ceiling grows sublinearly in its
number of \emph{value samples},
\begin{equation}
|\mathrm{children}(n)| \;<\; \bigl\lceil c_w \, m(n)^{\alpha} \bigr\rceil,
\qquad 0 < \alpha < 1
\label{eq:widen}
\end{equation}
which is standard progressive widening \cite{coulom2007,chaslot2008}, a device for search spaces with
unlimited possible actions. Keying \eqref{eq:widen} on $m$ rather than on $N$ has a consequence worth
stating: a stream of rejected mutations raises $N$ but not $m$, so it can never reopen a node for
further offspring. A node earns more children only by producing scored ones.

\begin{figure}[t]
\hrule\vspace{2pt}
\noindent\textbf{Algorithm 1: BONSAI}
\vspace{1pt}\hrule\vspace{3pt}
\small
\noindent\textbf{Input:} seed skill $s_0$, rollout budget $R$, splits train\,/\,val\,/\,test
\vspace{2pt}
\begin{algorithmic}[1]
\State root $\gets s_0$;\ \ $v(\text{root}) \gets \mathrm{Val}(s_0)$;\ \ seed the root's first layer
\While{rollouts $< R$}
    \State $n \gets$ descend from the root by \eqref{eq:uct} to a node under its cap \eqref{eq:widen}
    \State $B \gets$ next training minibatch; run the performer on $B$ under $\mathrm{skill}(n)$
    \State $c \gets$ optimizer rewrite of $\mathrm{skill}(n)$ from those scored attempts
    \If{$\sum_{i\in B} g_i(c) > \sum_{i\in B} g_i(n)$} \Comment{acceptance, Eq.~\eqref{eq:accept}}
        \State $v(c) \gets \mathrm{Val}(c)$; attach $c$ under $n$ \Comment{Eq.~\eqref{eq:backup}}
        \State back up $v(c)$ to every ancestor
    \Else
        \State back up a visit only to every ancestor \Comment{Eq.~\eqref{eq:backup-reject}}
    \EndIf
\EndWhile
\State \Return $\arg\max_n v(n)$, then evaluate the test split once
\end{algorithmic}
\vspace{2pt}\hrule
\end{figure}

\paragraph{Shipping:} When the budget is exhausted, BONSAI ships $n^{*} = \arg\max_{n} v(n)$, the
plain highest-fitness document, and evaluates the test split once on it. Evolvability decides only
where budget is spent, and shipping is deliberately kept separate. Any rule that discounted a document
by its brittleness would penalise the nodes the search probed most: a well-probed node
has a visible $\sigma$, whereas an unprobed leaf has $Q = v$ and no gap to charge, so such a rule would reward
ignorance. The search never reads the test set.

The tree the search maintains supports a second expansion operator beyond the single-document
rewrite, described next and evaluated in Table~\ref{tab:main}. A further per-lineage scratchpad that
gives mutation a memory is off by default and deferred to Appendix-A.

\subsection{GRAFT: Asymmetric Capability Transfer}

\begin{figure}[t]
\centering
\includegraphics[width=0.9\columnwidth]{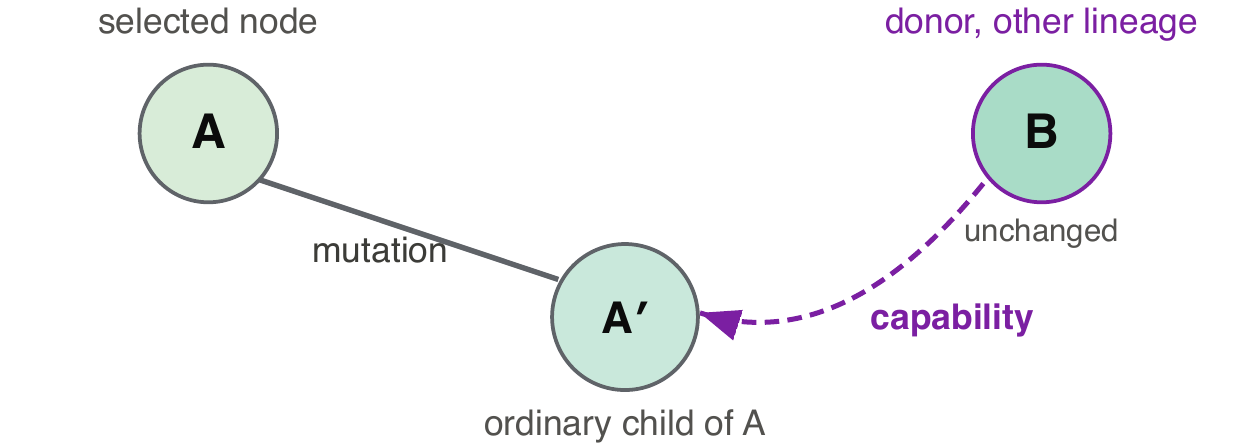}
\caption{The graft operator. The selected node $A$ is revised into an ordinary child $A'$ along the
solid edge, exactly as any other mutation would be. The cross-lineage donor $B$ contributes a
capability along the dashed edge, which carries evidence rather than ancestry: the donor gains no
visit and no value sample, and the search structure remains a tree.}
\label{fig:graft}
\end{figure}

Capability can end up divided across lineages: one branch of the tree learns to handle a family of
tasks that another branch continues to fail. No ordinary rewrite recovers the difference, because the
optimizer is shown only the selected document and a batch of its own failures, and it has no way of
knowing that the missing technique already exists elsewhere in the tree. The \textbf{graft} operator
supplies exactly that (Algorithm~2). The selected node $A$ acquires a capability that a cross-lineage node $B$
demonstrably possesses and that $A$ lacks. The operator is deliberately \emph{asymmetric}: rather than
combine two documents into a third, it produces a revision of $A$ that enters the tree as an ordinary
child of $A$, with the donor $B$ playing the role the reflection batch plays in any other expansion,
namely that of evidence rather than ancestry. The donor therefore receives neither a visit nor a value
sample, and every quantity defined above keeps its meaning (Figure~\ref{fig:graft}).

Grafting must earn its use. Every selection already scores a node on a training batch, so each node
accumulates, at no additional cost, a record $C_n$ of its per-task scores. Over the tasks on which $A$
and $B$ have both been scored, define the \textbf{payload} $B^{+} = \{\, i : C_B(i) > C_A(i) \,\}$,
the tasks on which $B$ outscores $A$, and the \textbf{guardrail} $A^{+}$ symmetrically, the tasks on
which $A$ outscores $B$ and must not be lost. Membership is a per-task score comparison rather than a
solved/unsolved cutoff, so partial credit transfers correctly: a donor that lifts a task's score
without fully solving it still contributes it. Grafting fires only when $|B^{+}| \ge \tau$
(Algorithm~2). This one-sided gate admits even a donor that dominates $A$ outright, the
most informative case, which a symmetric criterion would refuse. The optimizer is asked to restate the
donor's technique in $A$'s own terms. The shown set $S = B^{+} \cup A^{+} \cup D$ also carries a small
set $D$ of \textbf{anchors}, tasks on which the two already score alike, so the comparison is not taken
purely on the tasks that separate them. Every task shown to the optimizer is drawn from the training
split, so grafting reveals nothing about validation or test. The child is admitted only if
\begin{equation}
\sum_{S} g(c) \;>\; \sum_{S} g(A) \quad\text{and}\quad g_i(c) \ge g_i(A) \;\; \forall\, i \in A^{+}\cap S
\label{eq:graft-admit}
\end{equation}
The first condition is the acceptance rule \eqref{eq:accept}, so a grafted child faces the same test
every other child faces. The second makes the operator safe: because $A^{+}$ is by construction the
tasks on which $A$ outscores the donor, a child that merely reproduces the donor scores no higher than
the donor there, below $A$, and is rejected, so the tasks that make the donor an imperfect teacher are
precisely those that expose a copy. When $A^{+}$ is empty the donor dominates $A$ on the overlap and
the guardrail has nothing to test.
There, and only there, we additionally require $\sum_{S} g(c) > \sum_{S} g(B)$, so the child must add
something the donor did not already have.

\begin{figure}[t]
\hrule\vspace{2pt}
\noindent\textbf{Algorithm 2: GRAFT} (optional expansion operator)
\vspace{1pt}\hrule\vspace{3pt}
\small
\noindent\textbf{Input:} selected node $A$, per-node score records $C$, payload threshold $\tau$
\vspace{2pt}
\begin{algorithmic}[1]
\For{each cross-lineage node $B$ (neither ancestor nor descendant of $A$)}
    \State $B^{+} \gets \{\, i : C_B(i) > C_A(i) \,\}$ \Comment{payload}
    \State $A^{+} \gets \{\, i : C_A(i) > C_B(i) \,\}$ \Comment{guardrail}
\EndFor
\State $B^{*} \gets \arg\max_{B}\, |B^{+}|$ over all $B$ with $|B^{+}| \ge \tau$
\If{no such $B^{*}$ exists}
    \State fall back to the rewrite expansion (Algorithm~1); \textbf{return}
\EndIf
\State $S \gets B^{*+} \cup A^{+} \cup D$, \ with $D$ a set of anchors on which $A$ and $B^{*}$ score alike
\State show the optimizer $A$, the donor $B^{*}$, and the scored tasks in $S$
\State $c \gets$ optimizer revision of $A$ restating $B^{*}$'s technique in $A$'s terms
\State \textbf{admit} $c$ iff all three conditions hold: \Comment{admission, Eq.~\eqref{eq:graft-admit}}
\Statex \hskip1.4em (i)\ \ $\sum_{i \in S} g_i(c) > \sum_{i \in S} g_i(A)$
\Statex \hskip1.4em (ii)\ \ $g_i(c) \ge g_i(A)$ \ for every $i \in A^{+} \cap S$
\Statex \hskip1.4em (iii)\ \ $A^{+} \neq \emptyset$, \ or else $\sum_{i \in S} g_i(c) > \sum_{i \in S} g_i(B^{*})$
\If{$c$ is admitted}
    \State $v(c) \gets \mathrm{Val}(c)$; attach $c$ as an ordinary child of $A$ \Comment{Eq.~\eqref{eq:backup}}
    \State back up $v(c)$; the donor $B^{*}$ receives nothing
\Else
    \State back up a visit only \Comment{Eq.~\eqref{eq:backup-reject}}
\EndIf
\end{algorithmic}
\vspace{2pt}\hrule
\end{figure}

\section{Experiments}

\subsection{Setup}

We evaluate on three benchmarks. In \textbf{SpreadsheetBench} \cite{ma2024spreadsheetbench} each task
supplies a natural-language instruction and an input \texttt{.xlsx} workbook. The performer writes a
single Python script, which is executed in a sandbox, and the resulting workbook is compared cell by
cell against a golden output, making grading objective and automatic. The split is fixed at 80
train, 40 validation, and 280 test tasks. In \textbf{SearchQA} \cite{dunn2017searchqa} each item
supplies a quiz-style question together with the retrieved passages the performer may read. The
performer returns one short answer in a single attempt, scored by exact match after the standard
normalisation. The split is fixed at 400 train, 200 validation, and 1400 test questions. In
\textbf{LiveMathematicianBench} \cite{livemathbench} each item poses a mathematical statement with five
closely-worded candidate answers, exactly one of which is correct. The performer reads the statement
and its options and returns a single choice label, scored by exact match. The split is 60 train, 60
validation, and 57 test items. Splits are identical across every run and method, and on all three
benchmarks the objective is exact match alone. Partial credit is computed and logged but never steers
the search.

The performer is granite-4.1-30b, frozen, at temperature 0, and the optimizer is DeepSeek-V3.2 at
temperature 0.7 throughout. Within each benchmark both arms use the same performer and optimizer. A
run is allotted roughly 2400 performer rollouts on SpreadsheetBench, 18{,}000 on SearchQA, and 3000 on
LiveMathematicianBench, where a rollout is one attempt by the performer at one task, and every reported search
result is a single run at seed 42. The remaining constants are $\lambda = 1$, $c_w = 1$, and
$\alpha = 1/2$, with a first layer of 4 children. The exploration constant $c$ and the reflection
batch size are stated per run in Table~\ref{tab:main}.

\paragraph{Baselines:} We compare against \textbf{GEPA} \cite{agrawal2025gepa}, a reflective
prompt-evolution method that keeps a Pareto frontier of candidates, a set in which each member is best
on at least one validation task, and mutates a member sampled from it. GEPA's own machinery is left
untouched, and only the shared inputs are matched: the same performer, optimizer, seed document,
splits, and scorer, and a budget capped at BONSAI's \emph{measured} rollout count, with both arms
counting only uncached performer calls so a rollout means the same in each. We add a second
budget-matched baseline, \textbf{SkillOpt} \cite{skillopt2026}, which treats the skill as trainable
state. It reflects on minibatches of trajectories to propose edits, the same minibatch structure GEPA
and BONSAI use, but then routes those edits through an \emph{evidence-blind} aggregation: a
hierarchical merge reconciles and deduplicates them and a ranking step keeps the top edits under a
learning-rate budget, both stages seeing the edits and their written justifications but not the
trajectories that produced them. We match its reflection minibatch to ours and hold the same performer,
optimizer, seed, splits, scorer, and measured budget. We also evaluate the frozen
performer on the same test split with \emph{no skill at all}, and with the \emph{seed skill} alone,
the document SkillOpt \cite{skillopt2026} uses verbatim and every search arm starts from.

\subsection{Results}

\begin{table}[t]
\centering
\small
\setlength{\tabcolsep}{4pt}
\begin{tabular*}{\columnwidth}{@{\extracolsep{\fill}}>{\raggedright\arraybackslash}p{0.32\columnwidth}c>{\raggedleft\arraybackslash}p{0.32\columnwidth}@{}}
\toprule
\textbf{Method} & \textbf{Accuracy (\%)} & \textbf{Solved} \\
\midrule
\multicolumn{3}{l}{\emph{SpreadsheetBench} (batch 5, $c=0.4$, ${\sim}2400$ rollouts)} \\
No skill              & 7.50  & 21 / 280 \\
Seed skill            & 17.50~\dg{10.00} & 49 / 280 \\
GEPA \cite{agrawal2025gepa} & 21.07~\dg{13.57} & 59 / 280 \\
SkillOpt \cite{skillopt2026} & 20.00~\dg{12.50} & 56 / 280 \\
BONSAI (ours) & 23.21~\dg{15.71} & 65 / 280 \\
\textbf{BONSAI + GRAFT} & $\mathbf{25.00}$~\dg{17.50} & \textbf{70 / 280} \\
\midrule
\multicolumn{3}{l}{\emph{SearchQA} (batch 8, $c=0.5$, ${\sim}18{,}000$ rollouts)} \\
No skill              & 72.50 & 1015 / 1400 \\
Seed skill            & 72.43~\dr{0.07} & 1014 / 1400 \\
GEPA \cite{agrawal2025gepa} & 78.29~\dg{5.79} & 1096 / 1400 \\
SkillOpt \cite{skillopt2026} & 75.57~\dg{3.07} & 1058 / 1400 \\
\textbf{BONSAI (ours)} & $\mathbf{79.00}$~\dg{6.50} & \textbf{1106 / 1400} \\
BONSAI + GRAFT & 78.93~\dg{6.43} & 1105 / 1400 \\
\midrule
\multicolumn{3}{l}{\emph{LiveMathematicianBench} (batch 8, $c=1.25$, ${\sim}3000$ rollouts)} \\
No skill              & 17.74 & 10 / 57 \\
Seed skill            & 28.23~\dg{10.49} & 16 / 57 \\
GEPA \cite{agrawal2025gepa} & 56.14~\dg{38.40} & 32 / 57 \\
SkillOpt \cite{skillopt2026} & 59.65~\dg{41.91} & 34 / 57 \\
\textbf{BONSAI (ours)} & $\mathbf{64.91}$~\dg{47.17} & \textbf{37 / 57} \\
BONSAI + GRAFT & 63.16~\dg{45.42} & 36 / 57 \\
\bottomrule
\end{tabular*}
\caption{Held-out accuracy on all three benchmarks. Each search arm is a single run at seed 42 with a
frozen granite-4.1-30b performer and a DeepSeek-V3.2 optimizer, and GEPA is capped at BONSAI's measured
rollout count. The no-skill and seed-skill rows are single evaluations of the frozen performer on the
same test split. The small coloured figure after each accuracy is its change over the no-skill row,
green for a gain and red for a loss.}
\label{tab:main}
\end{table}

On SpreadsheetBench, our primary benchmark, BONSAI improves on the document it starts from by $5.71$
accuracy points and on GEPA by $2.14$. The skill-free rows set the scale:
writing the seed by hand is worth $10.00$ points over giving the agent no instruction at all, and
searching onward from it adds a further $5.71$. The ordering repeats on the other two benchmarks. On
SearchQA the seed is a bare stub rather than a field manual and a skill-free agent already answers
$72.50$\% of the questions, so essentially the whole of BONSAI's $6.57$-point gain over the seed is
attributable to the search. LiveMathematicianBench is where the gap is widest: BONSAI reaches $64.91$\% against a
seed of $28.23$\% and a skill-free $17.74$\%, clearing GEPA by $8.77$ points. Within
each benchmark both search arms share the same mutation operator, acceptance rule, performer,
optimizer, seed document and budget, and differ only in how they organise the search, so the margin
between them is attributable to the search strategy rather than to the edits it makes.

SkillOpt, the second budget-matched baseline, reaches $20.00$, $75.57$, and $59.65$: on average level
with GEPA ($51.74$ against $51.83$) and $3.97$ points below BONSAI. Its shortfall is concentrated in
aggregation rather than reflection. The per-minibatch analyst edits are useful, but the merge and rank
stages compress them sharply, a single step routinely dropping from seventeen candidate edits to
three, and most of the resulting candidates fail the validation gate. Because those stages reconcile
and rank edits without seeing the trajectories that motivated them, the method depends on the
optimizer's ability to judge edits in the abstract more than the single grounded rewrite GEPA and
BONSAI use, and that dependence surfaces once the optimizer is held fixed at DeepSeek-V3.2 across every
arm rather than supplied by a stronger model.

Table~\ref{tab:main} also reports \textbf{BONSAI + GRAFT}, the search run with the graft operator of
Section~2.4 enabled. A graft fires only when a cross-lineage donor outscores the selected node on
training tasks it fails, so the operator is self-limiting: where lineages converge on nearly the same
tasks it rarely triggers and defaults to the ordinary rewrite. This is what SearchQA and
LiveMathematicianBench show, benchmarks on which the two arms differ only in whether grafting is
enabled: the operator leaves the result within a single held-out task of the base method ($78.93$\% and
$63.16$\%). On SpreadsheetBench, where lineages differentiate more, grafting is active throughout the
run and the shipped skill reaches $25.00$\%.

\subsection{Where the budget goes}

Figure~\ref{fig:tree} shows the complete SpreadsheetBench run. Budget is unevenly allocated: the five
most-visited nodes below the root take 58 percent of all visits, and the tree reaches depth 4 across
32 documents, deep rather than wide, as selection on $Q$ should produce. Brittleness stays low, with
27 of the 32 nodes at $\sigma \le 0$ and the range spanning $-0.069$ to $+0.089$. The acceptance test
\eqref{eq:accept} is selective, admitting 31 of 131 proposals, so rejected proposals redirect the
search rather than waste it. Performer rollouts are the budget that matters. Optimizer calls number
131, and evolvability adds to neither, since \eqref{eq:Q} reuses scores already paid for.

\subsection{The evolvability signal}
\label{sec:lambda}

The comparison against GEPA differs in more than the selection signal. To isolate the contribution of
evolvability itself, we rerun the \emph{same} search with $\lambda = 0$, so the exploitation term is
raw fitness $v$ rather than the evolvability value $Q$ \eqref{eq:uct}. This is greedy score-chasing on
an identical tree, with the identical acceptance test \eqref{eq:accept} and the identical $\arg\max v$
ship. Only the selection signal changes. On all three benchmarks, steering by evolvability wins on held-out
test (Table~\ref{tab:lambda}).

\begin{table}[t]
\centering
\small
\setlength{\tabcolsep}{6pt}
\begin{tabular}{lccc}
\toprule
\textbf{Benchmark} & \textbf{$\lambda{=}1$ ($Q$)} & \textbf{$\lambda{=}0$ ($v$)} & \textbf{$\Delta$} \\
\midrule
SpreadsheetBench & $\mathbf{23.21}$ & 20.00 & $+3.21$ \\
SearchQA & $\mathbf{79.00}$ & 76.86 & $+2.14$ \\
LiveMathematicianBench & $\mathbf{64.91}$ & 57.89 & $+7.02$ \\
\bottomrule
\end{tabular}
\caption{Held-out test accuracy (\%) under evolvability selection ($\lambda{=}1$, the exploitation
term is $Q$) versus greedy selection ($\lambda{=}0$, the term is raw $v$). Same tree, same acceptance
rule, same $\arg\max v$ ship. Only the selection signal differs.}
\label{tab:lambda}
\end{table}

The mechanism behind the gap is visible in how far each search climbs. Greedy selection reaches a
validation peak early and then stalls, spending the rest of the budget on a region that no longer
improves, while evolvability keeps discovering higher-scoring documents deeper into the run. On
SearchQA the greedy run's best validation score stops rising at iteration 13 ($0.760$) while the
evolvability run climbs to $0.775$ by iteration 23. The other two benchmarks show the same split: on
LiveMathematicianBench, $0.700$ (greedy, iteration 16) against $0.750$ (evolvability, iteration 43),
and on SpreadsheetBench, $0.200$ (iteration 20) against $0.275$ (iteration 95). Because both arms ship
$\arg\max v$ and share every other component, the held-out gap is attributable to the selection signal
alone.

\subsection{Limitations}

The evidence is a single run per benchmark with one performer and optimizer pairing. Because BONSAI
and the baseline differ both in tree structure and in the selection signal ($\lambda = 1$ on $Q$), the
reported gain against the baseline reflects the search strategy as a whole. Section~\ref{sec:lambda}
isolates the evolvability component on its own, holding the tree fixed. The scratchpad (Appendix-A) is
described but not evaluated here. Finally, the acceptance test
\eqref{eq:accept} is decided on a small batch of five to eight tasks, so its reliability bounds what
any search built on it can achieve, a limitation shared with the baseline and one that binds hardest
where a small batch discriminates weakly, as it does on SearchQA.

\section{Related Work}

Instruction optimisation has been approached by search over paraphrases and edits
\cite{zhou2023ape,yang2024opro}, by iterative self-revision \cite{madaan2023selfrefine}, and by
compiling modular programs \cite{khattab2024dspy}. Two systems sit closest to ours. \textbf{GEPA}
\cite{agrawal2025gepa} performs the same reflective mutation, but its Pareto frontier is defined with
respect to the finite validation tasks. A prompt is retained because it is nondominated on that
particular set of instances, and later reflections explicitly combine complementary lessons from
prompts that are each strong on different validation instances. Its notion of diversity is therefore
\textbf{instance diversity}: a candidate survives because it is exceptional on some subset of the
validation tasks. Our notion of diversity is completely different. A node survives because its neighbourhood
remains productive under mutation. Those are orthogonal notions. A Pareto frontier over validation
instances encourages preservation of prompts that explain idiosyncratic variation in the sampled
validation set. BONSAI instead allocates budget according to \textbf{mutational robustness}, a
property of the optimisation landscape rather than of a finite validation sample. \textbf{SkillOpt} \cite{skillopt2026} casts the skill as trainable state and advances a
single trajectory under a learning-rate cap and a validation gate, so it cannot reconsider a region it
has left. Its update also splits reflection from composition: the optimizer merges and ranks the
reflected edits under an edit budget while seeing the edits but not the trajectories behind them, so
the composition step leans on optimizer strength where our single grounded rewrite does not. Both
decide on a point estimate. BONSAI can also return to an earlier region when a later one
stops paying.

\section{Conclusion}

BONSAI turns skill optimisation into an evolvability-guided search: by growing every child as a
mutation and descending with the upper-confidence rule~\eqref{eq:uct}, whose exploitation term weighs
a region's evolvability against a skill's own fitness, it steers budget toward the regions that keep
improving. On SpreadsheetBench, at an equal budget, it beats the strongest budget-matched baseline by
$2.14$ held-out points and the document it starts from by $5.71$, at no added cost.

\bibliography{bonsai}

\clearpage
\section*{Appendix-A\\A Lineage-Scratchpad}

The reflective rewrite (Section~2.3) and the GRAFT operator (Section~2.4) both act by proposing a new
document. The tree the search maintains supports one further mechanism that neither uses: a per-lineage
\textbf{scratchpad} that gives mutation a memory. It is off by default and the experiments in the main
text do not use it; we describe it here for completeness.

Every rewrite the optimizer proposes is judged, and the outcome is worth keeping whichever way it falls. After each attempt, the optimizer is asked to name, in two or three short lines, exactly what it changed. That note is filed at the version it revised: as a success if the rewrite was kept and its held-out score exceeded that version, and as a \textbf{did-not-work} entry otherwise. Successes are recorded but never shown back: the change they describe is already written into the document being read, and surfacing it besides would risk making the optimizer \textbf{myopic}, anchoring its rewrites on repeating what has already worked rather than exploring what has not yet been tried. Only did-not-work entries are ever resurfaced, so the optimizer is not steered toward proposing the same change twice under a different phrasing. Should an entry once filed as unsuccessful later prove otherwise, tried again at the same version and this time kept, it is withdrawn, since it would otherwise assert something a later attempt has disproved.

These entries compose along the tree. When a version is chosen for its next rewrite, BONSAI gathers the did-not-work entries filed at that version and at every version on the path back to the root, oldest first, deduplicated, keeping only the most recent when the path is long. Figure~\ref{fig:scratchpad} shows this for a version $S8$ reached along $S1 \to S3 \to S8$. The entry filed at $S1$ comes from a rewrite that failed to improve on it, shown here as the dashed sibling beneath $S1$ that never entered the tree, and the same happens at $S3$. When $S8$ is later selected, its own record and both ancestors' are combined into the list the optimizer is shown, so a change already tried and found wanting anywhere along this path is not proposed again.

This concatenation is the version's \textbf{lineage-scratchpad}, assembled fresh at the moment it is needed rather than kept as a separate memory store. What keeps it from growing without bound is the \textbf{Observer}: once a version's own did-not-work list has taken five new entries, the optimizer itself, now in a curator role and spending its own calls rather than performer rollouts, is asked to compress that version's list in place, merging near-duplicates, dropping entries a later one has superseded, and holding it to six. Because compaction runs on each version independently, a version many rewrites deep still hands the optimizer a short, current list rather than the accumulated history of its whole lineage. Neither model is ever told it is exploring a tree: an entry is phrased only as a change already tried on this instruction and the outcome it had, so the vocabulary of the search itself never leaks into a prompt.

\par\medskip
\noindent\centerline{\includegraphics[width=0.9\columnwidth]{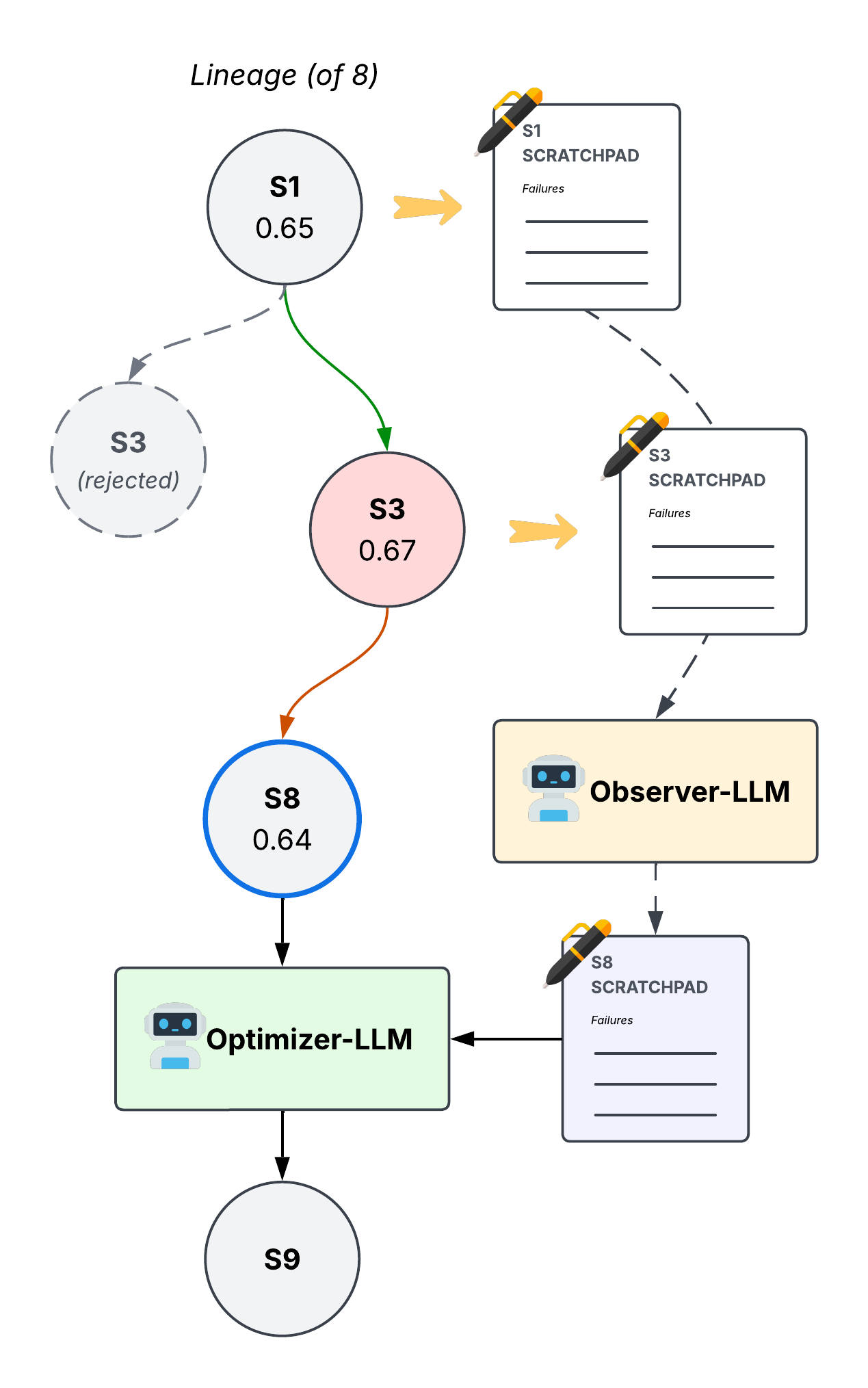}}
\vspace{3pt}
\captionof{figure}{Assembling the did-not-work record for a version $S8$ reached along $S1 \to S3 \to S8$. The entries already filed at ancestors $S1$ and $S3$, such as the rejected sibling proposal beneath $S1$, are gathered with $S8$'s own into the list the optimizer conditions on together with $S8$ itself to propose the next child, $S9$. Each ancestor's list is kept short by the Observer, which compacts a version's own record independently once enough entries accumulate on it.}
\label{fig:scratchpad}
\par\medskip

\section*{Appendix-B\\Search Trees Across Benchmarks}

\begin{figure*}[p]
\centering
\includegraphics[width=\textwidth]{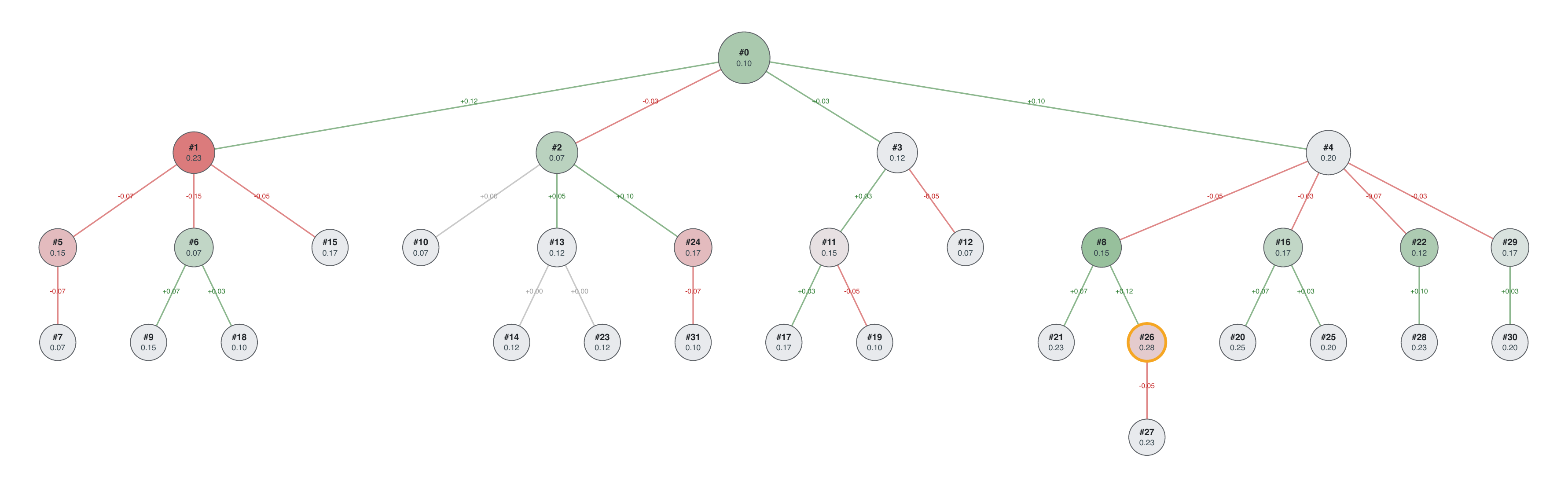}\\[8pt]
\includegraphics[width=\textwidth]{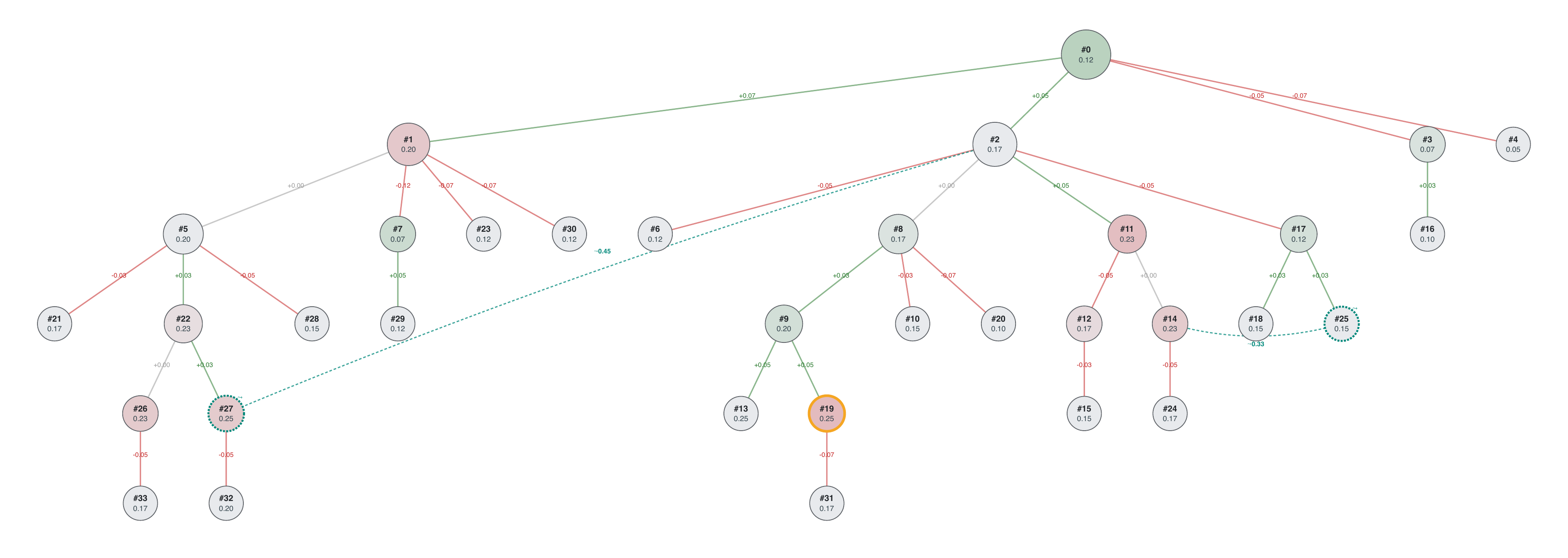}
\caption{\textbf{SpreadsheetBench} search trees, drawn with the encoding of Figure~\ref{fig:tree}
(fill = brittleness $\sigma$, circle area = value samples $m(n)$, edge labels = $\Delta v$, gold ring
= the shipped skill). \emph{Top:} BONSAI grows 32 documents to depth 4 and ships \#26 ($23.21$\%).
\emph{Bottom:} BONSAI\,+\,GRAFT grows 34 documents to depth 5 and ships \#19 ($25.00$\%); the two teal
dashed arcs mark the admitted grafts (\#25 and \#27), each an ordinary child of its selected node that
also draws a capability from a cross-lineage donor.}
\label{fig:trees-sb}
\end{figure*}

\begin{figure*}[p]
\centering
\includegraphics[width=\textwidth]{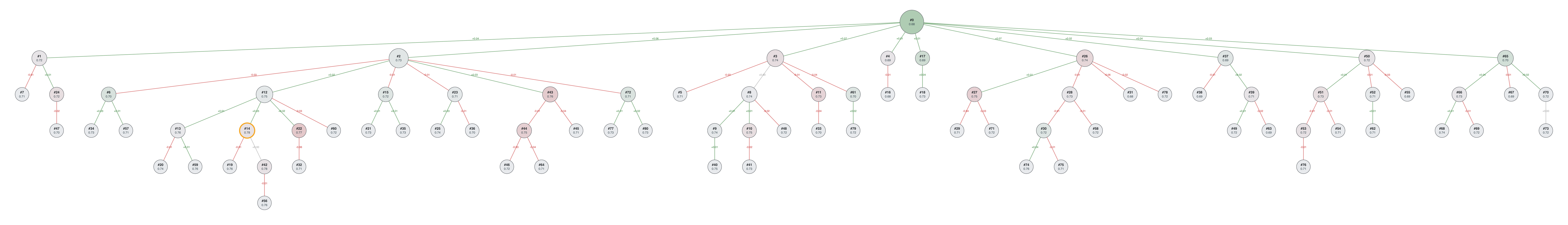}\\[8pt]
\includegraphics[width=\textwidth]{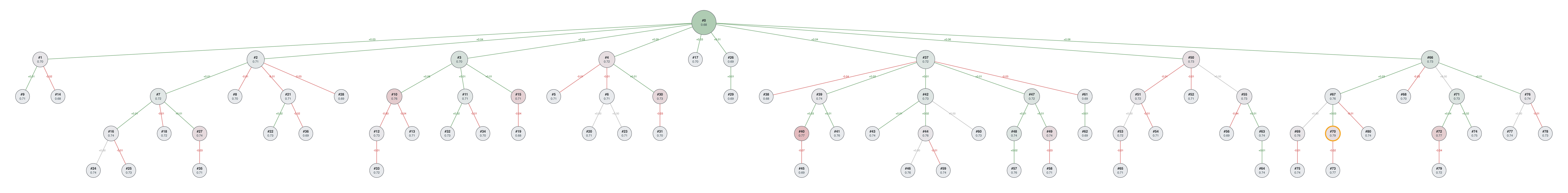}
\caption{\textbf{SearchQA} search trees (encoding of Figure~\ref{fig:tree}), the largest of the three
at $18{,}000$ rollouts and $81$ documents each. \emph{Top:} BONSAI ships \#14 ($79.00$\%).
\emph{Bottom:} BONSAI\,+\,GRAFT ships \#70 ($78.93$\%). No graft was admitted here: SearchQA lineages
converge on nearly the same tasks and rarely present the cross-lineage payload a graft needs, so the
tree is structurally indistinguishable from the rewrite-only run.}
\label{fig:trees-sqa}
\end{figure*}

\begin{figure*}[p]
\centering
\includegraphics[width=\textwidth]{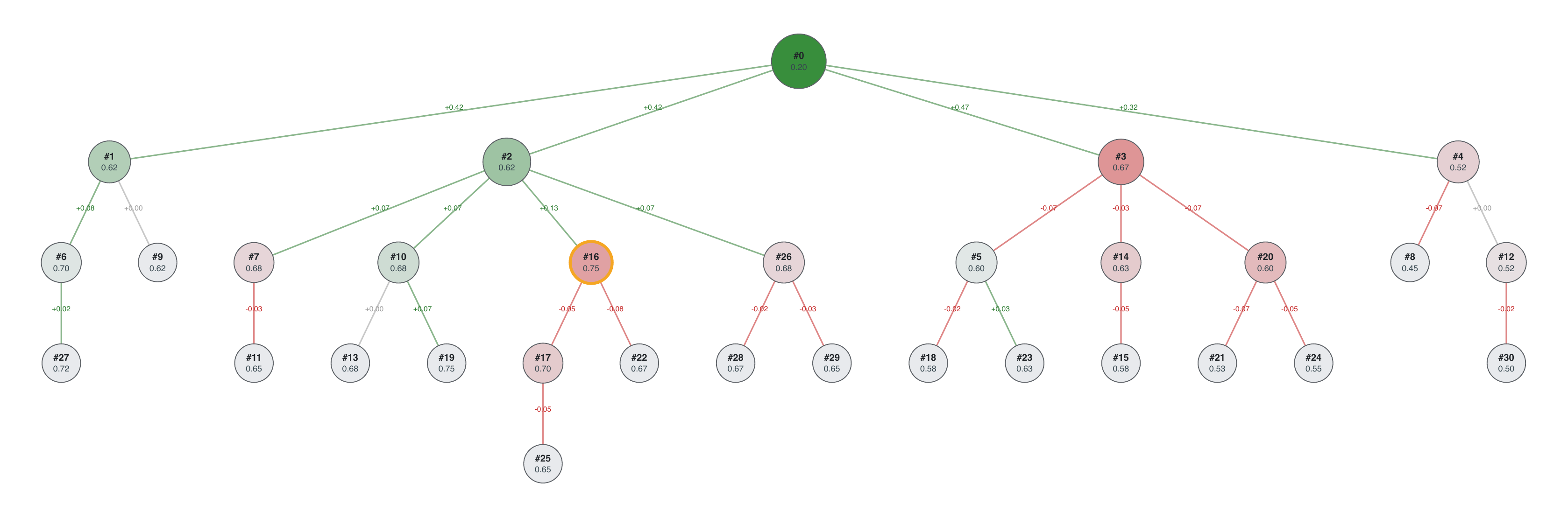}\\[8pt]
\includegraphics[width=\textwidth]{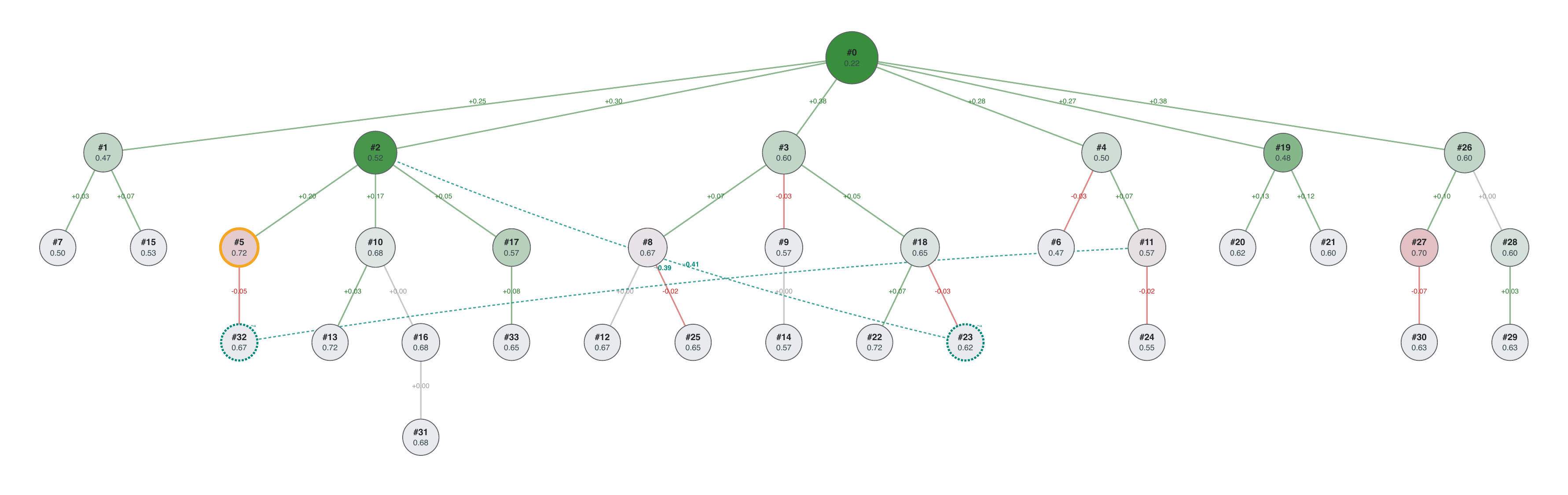}
\caption{\textbf{LiveMathematicianBench} search trees (encoding of Figure~\ref{fig:tree}).
\emph{Top:} BONSAI grows 31 documents to depth 4 and ships \#16 ($64.91$\%). \emph{Bottom:}
BONSAI\,+\,GRAFT grows 34 documents and ships \#5 ($63.16$\%); the two teal dashed arcs mark the
admitted grafts (\#23 and \#32), neither of which became the shipped skill.}
\label{fig:trees-lm}
\end{figure*}

For completeness, Figures~\ref{fig:trees-sb}--\ref{fig:trees-lm} show the full search trees behind every
run in Table~\ref{tab:main}: for each benchmark, the base BONSAI run and the BONSAI\,+\,GRAFT run,
drawn with the same encoding as the main-text tree (Figure~\ref{fig:tree}). They make the search's
shape visible, how deep each run drives its most evolvable lineages, how unevenly budget is allocated,
and where a graft entered the tree.

\end{document}